# Consideration of Vehicle Characteristics on the Motion Planner Algorithm

Syed Adil Ahmed and Taehyun Shim*

*University of Michigan Dearborn, Dearborn, MI 8030, USA*

(e-mail: adilsa@umich.edu & tshim@umich.edu). *Corresponding author

**Abstract**: Autonomous vehicle control is generally divided in two main areas; trajectory planning and tracking. Currently, the trajectory planning is mostly done by particle or kinematic model-based optimization controllers. The output of these planners, since they do not consider CG height and its effects, is not unique for different vehicle types, especially for high CG vehicles. As a result, the tracking controller may have to work hard to avoid vehicle handling and comfort constraints while trying to realize these sub-optimal trajectories. This paper tries to address this problem by considering a planner with simplified double track model with estimation of lateral and roll based load transfer using steady state equations and a simplified tire model to reduce solver workload. The developed planner is compared with the widely used particle and kinematic model planners in collision avoidance scenarios in both high and low acceleration conditions and with different vehicle heights.

## 1. INTRODUCTION

Research on autonomous vehicle (AV) control is an active research topic. In this area, motion planning for safe and efficient vehicle path is a necessity step for level 4 or above autonomous vehicles. Motion planning has been studied extensively for autonomous vehicles and a lot of methods exists for tackling this problem (survey paper (Paden et al., 2016)). These methods include sampling methods like A* and RRT* (Karaman & Frazzoli, 2011), vehicle velocity and path decoupling (Kant & Zucker, 1986; Shi et al., 2021) and optimization function-based methods (MPC, polynomial functions and others) (Falcone et al., 2008; Weiskircher et al., 2017). Amongst all of these methods the optimization-based methods, and in particular the Model Predictive Control (MPC), provides the most guarantees of optimal solution and also the ability to place physical and actuator constraints to the problem. Hence, methods using MPC for trajectory planning are considered in this paper.

Amongst the methodologies for AV control, the most used method involves a hierarchy-based control scheme. Here, the planning is an upper-level control and tracking is the lower level control. Fig. 1 shows a general flow chart view of the control scheme. In this scheme, the planner constitutes mostly a simple model. There are particle models used in (Falcone et al., 2008; Weiskircher et al., 2017). Here constraints are added for acceleration using the friction circle and limited by $\mu g$, and the cascaded actuator limit constraint on the control inputs of acceleration and yaw rate. In (Weiskircher et al., 2017) it is suggested that the particle-based planner can be used for any range of acceleration, but, for less significant tracking errors it is suggested to consider $A_y \leq 4.5 m/s^2$. In addition, there are also kinematic planner models used, where again the same constraints as the particle planner are used. But these models differ from the particle model as they contain information of vehicle wheelbase and use Ackerman steering angle relations for yaw rate. As a result, they are used for path planning of large vehicles, specifically at low speeds (Oliveira et al., 2020), and for smaller vehicles (like sedan or SUV) (Polack et al., 2017), (Sun et al., 2022).

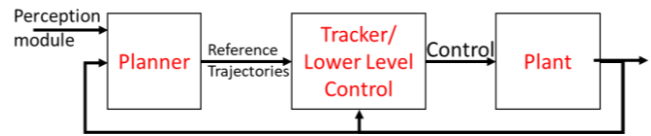

Figure 1: A hierarchical approach to AV control

For the hierarchical control scheme, the planner constitutes a simple model. This is because planners are tasked with a nonlinear task of finding an optimal path that avoids an obstacle, follows road curvature and at the same time also meet cascaded physical/actuator and environmental constraints. As a result, a simple model (having less states) allows a reduction in nonlinearity and a more likely real-time implementation, as the overall planner cost is smaller (Ferreau, 2011). Furthermore, the use of a simpler model in the planner is further made possible by use of a nonlinear higher order model in the tracker, which considers tire nonlinearity and mostly full car 7DOF+ models. These tracking controllers work at a faster rate (0.05s or lower) than the planner and ensure that a realistic actuator output meeting the actuator constraints is delivered, while also trying to follow the planner trajectory.

But, by using a simple planar particle model (3DOF) in the motion planner, we compromise vehicle trajectory tracking, handling and driver comfort. This is because the planner generates the same reference trajectories for different types of vehicles, which would imply that a minivan or a sedan would be expected to carry out a maneuver (for example, double lane change or taking a sharp curve) in the same way. At lower speeds, it may be possible that the same reference trajectories are used for both vehicles. But, at high speeds or emergency maneuvers, a sedan, having a very low CG (center of gravity) position, can probably use the trajectory outputted by the particle model-based planner, but a minivan that has a higher CG, mass, and inertia would be prone to large load transfers, roll and discomfort. Therefore, in order to output unique trajectories for different vehicle types, there is a need for adding some form of higher order modelling/modifications, especially at higher accelerations and limit maneuvers. This use of higher-order models in the planner for limit maneuvers is also recommended by Polack et al. in (Altché, 2017), where a comparison between kinematic and dynamic (9DOF) based planners is done and it is concluded that the former is only valid till $0.5\mu g$ lateral acceleration limits.

The use of higher order model as a planner could be one way of addressing the problem highlighted above. Another approach is the amalgamation of the planner and tracker to create an integrated system (Laurense & Gerdes, 2022) (Fig. 2), which uses a high order model (such as a bicycle or a double track model) in the integrated system. This approach claims that the hierarchical AV control (Fig. 1) results in conservative references from the planner due to constraints used to reduce infeasibility of planner reference, for example steering constraint for rollover protection. But the integration of the planner and tracker results in the requirement that the single controller needs a fast update frequency to work with vehicle actuators, while also including nonlinearity of vehicle model and tire model and satisfying the mostly non-convex collision avoidance constraints. These are competing objectives, which require a compromise, which is mostly done on the modeling side. In (Laurense & Gerdes, 2022) the authors use an integrated planner and tracker with a cascaded bicycle and point mass model for aggressive emergency maneuvers. The cascaded model is required since without it the real-time implementation of the controller becomes an issue, and the prediction horizon needs to be compromised (reduced). To the authors knowledge, no real-time integrated planner and tracker with nonlinear double track model, (which considers lateral and roll based load transfer), exists, and it is thought that this is because of the large computation time required for such a controller.

In this work we will use the AV control scheme highlighted in Fig. 1, as it allows more freedom to modify vehicle models without a penalty for the real-time implementation. Using this scheme, we compare our newly developed planner and reference planners (kinematic and particle) to show that the reference trajectories from motion planners need to consider the vehicle characteristics (such as size), particularly for a larger vehicle, like an SUV. Our new planner considers a novel double track model with simplifications of steady-state lateral and roll load transfer, a simplified tire model, and the removal of wheel speed states. This model is also compared with an 8DOF model to show its accuracy, and this modification allows the planner to be real-time compatible and output more optimal reference trajectories. The new planner is compared with the reference planners in two collision avoidance scenarios to show that trajectories differ between the reference and the new planners.

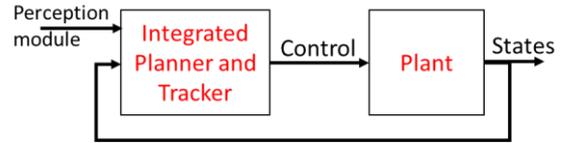

Figure 2: Integrated planner and tracker for AV Control

In this work, the reference planner model is introduced in section 2. In section 3, the new planner model is shown together with its comparison to the 8DOF model. Then in section 4 the implementation of the new planner is highlighted. Finally, in section 5 and 6, the analysis of the simulated results and a conclusion is provided, respectively.

## 2. REFERENCE PLANNER MODELS

The reference planner is based on a particle model or a kinematic bicycle model with yaw rate and steering respectively being one of the inputs and acceleration being the other input for both models. Both models can be built on the Frenet coordinate system, allowing easier curve handling.

*2.1 Particle Model*

Like the particle model in (Falcone et al., 2008; Weiskircher et al., 2017), this particle model is constructed with yaw rate, $\dot{\psi}_p$ and acceleration, $a_t$, being inputs to it. The model, (1), also consists of 6 states, where $v_t$ is the tangential velocity, $\psi_e$ is the yaw error between particle and reference yaw, $y_e$ is the lateral position error between vehicle and reference and $s$ is the distance traveled by particle along the reference path. The last two states are used for enforcing collision constraints, with $x_t$ being time and $\zeta_{CA}$ is the slack variable for collision avoidance constraint enforcement.

$$\dot{x} = \frac{d}{dt}\begin{bmatrix} v_t \\ \psi_e \\ y_e \\ s \\ x_t \\ \zeta_{CA} \end{bmatrix} = \begin{bmatrix} a_t \\ \dot{\psi}_p - v_t \cos(\psi_e)\left(\frac{\kappa}{1-y_e\kappa}\right) \\ (v_t)\sin(\psi_e) \\ v_t \cos(\psi_e)\left(\frac{1}{1-y_e\kappa}\right) \\ 1 \\ u_{CA} \end{bmatrix} \quad (1)$$

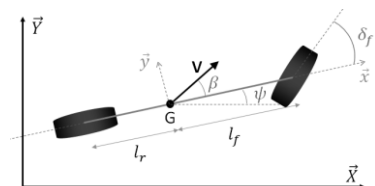

Figure 3: Kinematic bicycle model

*2.2 Kinematic model*

The kinematic bicycle model, Fig. 3, for the planner considers the vehicle wheelbase and its effect on yaw. And due to the reference point being the CG and not the rear axle, the side slip angle, $\beta$, also needs to be considered for yaw rate calculation. Also, the inputs to the model are, $A$, acceleration and $\delta$, steering. Otherwise, the model is like the particle model.

## 3. NEW PLANNER MODEL DEVEOPMENT

$$\dot{x} = \frac{d}{dt}\begin{bmatrix} v \\ \psi_e \\ y_e \\ s \\ x_t \\ \beta \\ \delta \\ \dot{\psi} \\ \zeta_{CA} \end{bmatrix} = \begin{bmatrix} A \\ \dot{\psi} - v\cos\beta \cos(\psi_e)\left(\frac{\kappa}{1-y_e\kappa}\right) \\ v\cos\beta \sin(\psi_e) \\ v\cos\beta \cos(\psi_e)\left(\frac{1}{1-y_e\kappa}\right) \\ 1 \\ -\frac{l_r}{l_f+l_r}\sec(\delta)^2 \dot{\delta} \\ \dot{\delta} \\ \frac{v\tan\delta}{a+b}\cos\beta - \dot{\psi} \\ T_s \\ u_{CA} \end{bmatrix} \quad (2)$$

The reference planners, as shown in (1) and (2) do not consider vehicle inertias, and the size of vehicle, hence there is no differentiation between planning for large or small vehicles. Due to this condition, it is possible that the trajectory offered by this planner is not optimal for the vehicle. Hence leaving the tracking controller to work out a suboptimal path that doesn't follow the planned trajectory or get overworked.

In order to avoid these issues, we consider a simple form of a double track model with the simplified tire model. The model has inputs of steering, $\delta$ and total longitudinal force, $F_X$. Hence, removing the fast wheel speed states.

In order to calculate the vertical forces correctly, the planner model considers steady state roll moments for calculating lateral load transfer for front and rear, $\Delta F_{zf}$ and $\Delta F_{zr}$.

$$M_{\phi f} = K_{\phi f}\phi + \frac{mb}{l_f+l_r}a_y = \Delta F_{zf} t_f \quad (3)$$

$$M_{\phi r} = K_{\phi f}\phi + \frac{mb}{l_f+l_r}a_y = \Delta F_{zr} t_r \quad (4)$$

Using these load transfer terms we can create better estimates for our vertical force calculations.

$$F_{zfr,l} = \frac{mgl_r}{2(l_f+l_r)} - \frac{mh_{cg}}{2(l_f+l_r)}a_x \pm \Delta F_{zf} \quad (5a)$$

$$F_{zrr,l} = \frac{mga}{2(l_f+l_r)} + \frac{mh_{cg}}{2(l_f+l_r)}a_x \pm \Delta F_{zr} \quad (5b)$$

Where, $\phi = \frac{m(h_{cg}-h_{rc})}{K_{\phi f}+K_{\phi r}-W(h_{cg}-h_{rc})}a_y$, m is the mass, $K_{phi}$ is the suspension stiffness, $\phi$ is the roll angle, $t_f$ and $t_r$ are the front and rear wheelbase lengths, $a_x$ is the longitudinal acceleration and $a_y$ is the lateral acceleration.

The nonlinear differential equations governing this motion planner are as shown in (6). The model includes states of longitudinal velocity, $v_x$, lateral velocity, $v_y$, yaw rate, $\dot{\psi}$, and the curvilinear coordinates of $s, \psi_e, y_e$ and collision avoidance related constraints of $x_t$ and $\zeta_{CA}$.

In (6), $f_{yij}$, follows the simplified tire model provided in (Adireddy et al., 2010) and is shown in (7).

$$\dot{x} = \begin{bmatrix} m\dot{v}_x \\ \dot{\psi}_e \\ \dot{y}_e \\ \dot{s} \\ m\dot{v}_y \\ I_z\ddot{\psi} \\ \dot{x}_t \\ \dot{\zeta}_{CA} \end{bmatrix} = \begin{bmatrix} \frac{F_X F_{zf}}{mg}\cos\delta + \frac{F_X F_{zr}}{mg} - (f_{yfr}+f_{yfl})\sin\delta + mv_y\dot{\psi} \\ \dot{\psi} - (v_x \cos\psi_e - v_y\sin\psi_e)\left(\frac{\kappa}{1-y_e\kappa}\right) \\ v_x \sin\psi_e + v_y\cos\psi_e \\ (v_x\cos\psi_e - v_y\sin\psi_e)\left(\frac{1}{1-y_e\kappa}\right) \\ \frac{F_X F_{zf}}{mg}\sin\delta + f_{yrr} + f_{yrl} + (f_{yfr}+f_{yfl})\cos\delta - mv_x\dot{\psi} \\ (f_{yfr}+f_{yfl})a\cos\delta + \frac{(f_{yfl}-f_{yfr})t_f\sin\delta}{2} - (f_{yrr}+f_{yrl})b \\ 1 \\ u_{CA} \end{bmatrix} \quad (6)$$

$$f_{yij} = \begin{cases} C_\alpha \alpha_{ij} \; ; \; 0.85\alpha_0 \geq |\alpha| \\ \frac{C_\alpha}{6}(|\alpha_{ij}| + 4.25\alpha_0) sgn(\alpha_{ij}) \; ; 0.85\alpha_0 < |\alpha| < 1.75\alpha_0 \\ F_{peak} sgn(\alpha_{ij}) \; |\alpha| \geq 1.75\alpha_0 \end{cases} \quad (7)$$

Where, $C_\alpha = aF_{zij} + bF_{zij}^2$, $\alpha_0 = F_{peak}/C_\alpha$ and $F_{peak} = (0.9 - 0.182\left(\frac{F_{zij}}{F_{z0}} - 1\right))\mu F_{zij}$. Here, a and b are quadratic polynomial coefficients of cornering stiffness calculated from Magic Formula tire data for each vehicle. And $F_{z0}$ is the nominal tire force.

Fig. 4 shows that the model in (6) closely follows the nonlinear 8DOF model with Pacejka tire model. This check ensures that the output of states we get from the planner are realistic. Also, it allows us to later check other reference planners for trajectories (Y position, yaw rate and speed) that have conflicting references (like, some planners might have a Y position that has a very different yaw rate than what would be required by the vehicle). This is necessary, to limit the scope, as there is no tracking controller developed in this paper.

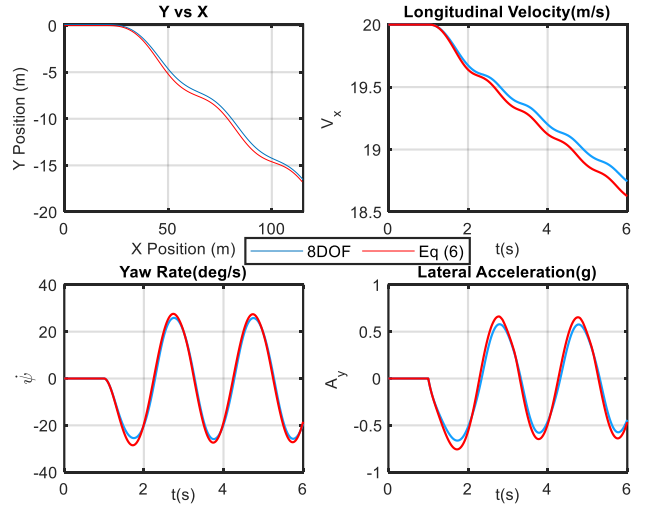

Figure 4: 8DOF vs (6) in 0.5Hz sine steer at initial speed of 20 m/s

## 4. IMPLEMENTATION OF THE NMPC PLANNER

The objective of a planner is to devise a path and motion profile that allows road curvature following, avoids obstacles, follows constraints (environment, speed limits and actuator) and provides a comfortable path. The following is the NMPC cost function and constraints:

$$J = \min_{x_k,u_k} \sum_{k=1}^{N_p+1} |y_k - y_{ref,k}|_Q^2 + \sum_{k=1}^{N_p} |\Delta u_k|_R^2 + \sum_{k=1}^{N_p} |u_k|_S^2 + J_\alpha \quad (8)$$

Where, $y = [y_e, v_x, v_x - \zeta_{CA}]$ and $u = [\delta, F_X, u_{CA}]$

s.t.

$$\dot{x}_k = f(x_k, u_k) \quad (8a)$$

$$y = Cx_k \quad (8b)$$

$$\delta_{min} < \delta < \delta_{max} \quad (8c)$$

$$-\frac{Max\ Engine\ Power}{v_x} < F_X < \frac{Max\ Engine\ Power}{v_x} \quad (8d)$$

$$\left(\frac{y_e - y_{e,o_i}}{\Delta w_{o_i}}\right)^2 + \left(\frac{s - s_{o_i}}{\Delta l_{o_i}}\right)^2 \geq 1 \quad (8e)$$

$$y_e \kappa < 1 \quad (8f)$$

$$\zeta_{CA} \geq 0 \quad (8g)$$

$$\left(f_{yir} + f_{yil}\right)^2 + \left(\frac{F_X F_{zi}}{mg}\right)^2 \leq \mu F_{zi}; i = f, r \quad (8h)$$

$$\dot{\delta}_{min} \leq \dot{\delta} \leq \dot{\delta}_{max} \quad (8i)$$

$$y_{e,min} < y_e < y_{e,max} \quad (8j)$$

In (8) the first term from left is for tracking of outputs, where lateral offset of zero and reference speed tracking is desired. The second term looks at reducing jerky input values by penalizing rate of change of steering $\Delta \delta$ and total longitudinal force, $\Delta F_X$. The third term aims to reduce the total output of control values to ensure an energy efficient and economical control. The rightmost term in (8), aims at penalizing excessive slip angle usage by the planner for all tires. This is because the tire model (7) saturates out as a constant value, and it is possible that the controller uses more than the required steering when above $1.75\alpha_0$.

$$J_\alpha = \begin{cases} \sum_{k=1}^{N_p+1} |\alpha_{ij,k} - \alpha_{0ij,k}|_T^2 & ; if\ \alpha_{ij,k} > \alpha_{0ij,k} \\ 0 & ; otherwise \end{cases} \quad (9)$$

*4.1 Collision Avoidance (CA) constraint*

The collision avoidance constraint is provided in (8e) and involves an ellipse based Euclidean norm (Weiskircher et al., 2017). During the prediction horizon, the objects (OV) in Frenet coordinates are assumed to travel at constant tangential and normal accelerations, start at an initial position $s_{o_i,0}$ & $y_{e,o_i,0}$ and speed $v_{t,o_i}$ & $v_{n,o_i}$. This gives the following object position in s and $y_e$ frame:

$$s_{o_i} = s_{o_i,0} + v_{t,o_i} x_t + \frac{1}{2} a_{t,o_i} x_t^2 \quad (10)$$

$$y_{e,o_i} = y_{e,o_i,0} + v_{n,o_i} x_t + \frac{1}{2} a_{n,o_i} x_t^2 \quad (11)$$

And then the main collision avoidance constraint is ellipse based Euclidean norm (8e). Where, for each obstacle, i, $\Delta l_{o,ss} = \sqrt{2\Delta l^2}$ and $\Delta w_o = \sqrt{2\Delta w^2}$ and $\Delta l_{o,i} = \Delta l_{o,ss} + f_{\zeta_{CA}} \zeta_{CA}$. Here, $f_{\zeta_{CA}}$ is a free variable and $f_{\zeta_{CA}} = T_s$. In fig. 5 $y_{e,buffer}$ is a lateral buffer which is kept as 0.15m in this work.

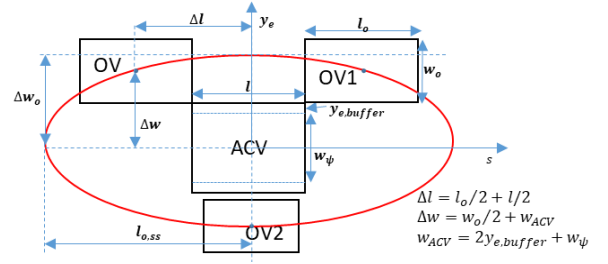

Figure 5: Ellipse based Euclidean norm

*4.2 Implementation*

Casadi 3.6.5 with MATLAB is used to solve the optimization problem. Casadi is a free and open-source symbolic framework for automatic differentiation and optimal control. The nonlinear optimization is solved using IPOPT solver, which is the default solver in Casadi. To speed up the solver, warm start procedure is used and multiple shooting discretization using Runge Kutta 4 method is used to reconstruct the nonlinear problem. The sample time ($T_s$) is set to 0.1s, while the prediction horizon is varied for each simulation.

## 5. RESULTS AND DISCUSSION

To evaluate the different motion planner models an obstacle avoidance scenario is created, that considers a static vehicle in ego lane and an oncoming traffic vehicle in opposite lane. The selected scenario causes the planners (reference and developed) to create their own required path around the obstacles.

For these scenarios, to see the effect of vehicle height, the developed controller is simulated with two versions of the SUV vehicle, h = $0.95 h_{cg}$ and h = $1.05 h_{cg}$ (loaded at top). Also, the tracking control is not used in this work, and the results will look to analyze planner outputs, specifically the planner outputs of yaw rate, vehicle position Y and velocity. These outputs form references for tracking control (like in (Adireddy et al., 2010)), and their optimality ensures that the tracking control is not overworked or follows an infeasible trajectory.

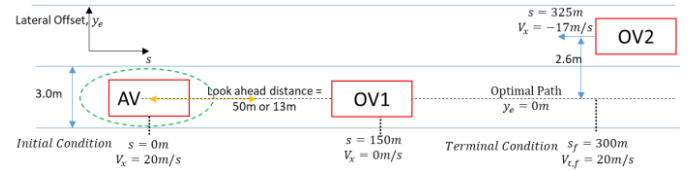

Figure 6: Collision avoidance scenario used to evaluate the planners

For the reference planners, a problem like (8) is solved, with the first three terms in (8). of the problem is slightly modified by considering constraints like (8c), (8d) and (8i) for the reference planner inputs and removal of constraint (8h). The sample time and prediction horizon are same as for the new planner.

The look ahead distance of 50m, $NV_xT_s$ (where N = 25), is selected for the controller, which results in linear tire region operation for the planners. To ensure linear region operation all planners have an added constraint of $A_y < 0.3g$.

Fig. 7 shows the state plot of the planners, while Fig. 8 shows that the maneuver is performed in a linear tire domain with a small amount of load transfer (LT). It is evident that all planners can navigate around the two obstacles in a timely fashion, albeit with small differences in the particle and kinematic planner outputs. The particle planner provides a much different reference for the tracking control, as it does not consider vehicle inertia or understeer gradient. The neutral steering kinematic planner has smaller steering and yaw rate compared to the new planner, as the SUV model is slightly understeering.

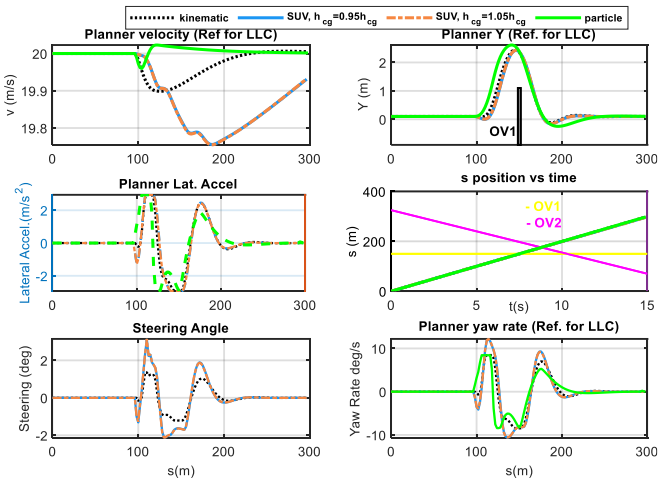

Figure 7: State outputs of planners in <0.3g conditions

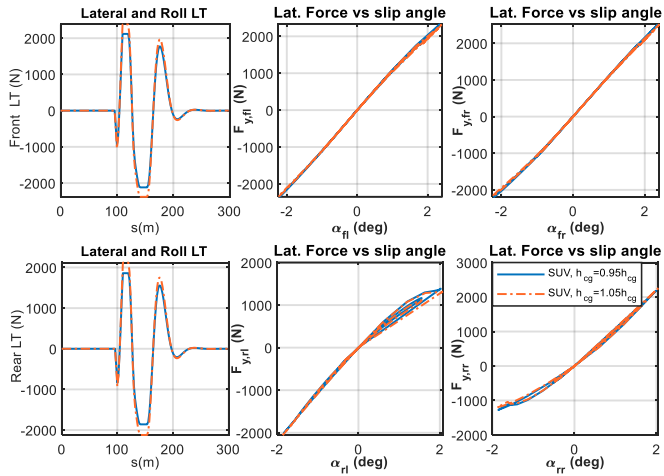

Figure 8: New planner load transfer and force vs slip angle plots

Also, in Fig. 7, it is seen that the references for the two SUV models are similar. In the linear region, the load transfer is very small and similar for the two SUV models as shown in Fig. 8.

Now, the look ahead distance of 26m (N=13) is selected for the controller, which corresponds to a sensor range malfunction and inability to see a static vehicle ahead till it is too late. Hence, this maneuver will check the nonlinear region operation of planners. The results for this case are shown in Figs. 9 and 10. In Fig. 9, we observe that the references for tracking controller, which include yaw rate, Y position, and velocity, are much different for the new planner and the kinematic and particle planners. The two reference planners output much less yaw rate than required for the maneuver of the large SUV vehicle, particularly at the nonlinear maneuver between 150-200m. This means that if these references are used in a tracking controller, it will have difficulty in following this tight trajectory. And perhaps this results in a suboptimal solution, as the traction available in the nonlinear region is much less than what is assumed by the kinematic and particle planners. Also, the particle and kinematic references differ from the new planner more in the nonlinear region operation in figs. 9 and 10, than in Figs. 7 and 8, showing the effects of vehicle height.

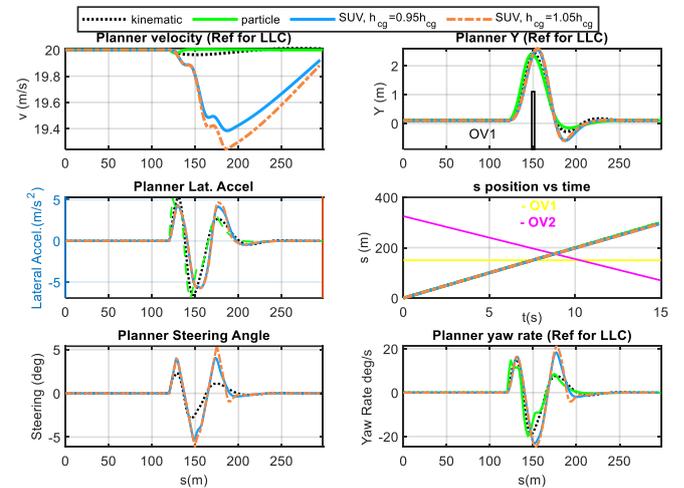

Figure 9: Planner state plots for nonlinear acceleration condition

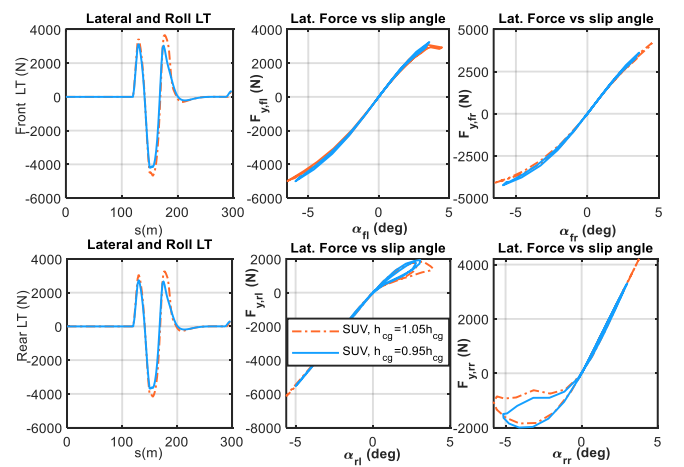

Figure 10: New planner load transfer and force vs slip angle plots

The reason for the difference in the reference and new planner is shown in Fig. 10, where it is observed that a considerable amount of lateral and roll LT is happening due to the vehicle's height. The significant LT causes the inside tires during the maneuver to be operate at their saturation limits, thereby, reducing the overall available lateral force and yaw moment to turn the vehicle. Also, shown in Figs. 9 and 10 is the difference between a loaded and nominal SUV. The top rack loaded SUV has more LT, resulting in the inside tires saturating. Therefore, the loaded SUV requires higher steering and yaw rates for the same path.

*5.1 Computation time.*

All the planners are simulated on a Windows 64-bit machine with a processing speed of 3.4 GHz, an i7 6700 CPU and 32 GB of RAM. The average simulation time for each planner is shown in Table 1. For N=13, all the computation times appear to be close to each other, and no extra penalty is accrued by using a higher-order planner model (6). Since code generation is not being utilized, it is not surprising that the times are slightly higher than the required sample time (0.1s). With code generation, it is expected that the results are going to be real-time compliant. But, for N>13, it may be necessary to consider some real-time implementation techniques that may involve specialized solvers like FORCES PRO, the use of control parametrization and blocking techniques, or real-time iteration schemes (Diehl et al., 2005).

Table 1: Average computation time for each planner

|  | Particle | Kinematic | New Planner |
|---|---|---|---|
| Avg. Computation time (s) at N=13 | 0.092 | 0.11 | 0.11 |
| Avg. Computation time (s) at N=25 | 0.11 | 0.13 | 0.16 |

## 6. CONCLUSIONS

In this work, a new planner model, which considers steady state roll and lateral load transfer, a simplified tire model, and a double track model is developed. The planner is proven to operate close to an 8DOF vehicle model and then simulated and compared along with the state-of-the-art planners in two collision avoidance scenarios. The work focuses on providing evidence that the simple models in the current state-of-the-art motion planners assume unlimited traction, hence providing undesirable references for the tracking controller. It is also shown that higher load transfer and tire nonlinear region operation results in the reference trajectories becoming more suboptimal and different from the newly developed planner.

In the future, longitudinal load transfer and its effects, along with the lateral effects on motion planning, will also be explored. Further, the addition of a tracking controller and its integration with the planners will be carried out to confirm the effects that the suboptimal trajectories have on the tracking controller.